\documentclass[journal]{IEEEtran}
\usepackage{cite}
\ifCLASSINFOpdf
   \usepackage[pdftex]{graphicx}
\else
\fi
\usepackage{subfig}
\usepackage{colortbl}
\usepackage{color}
\definecolor{grayblue}{rgb}{0.85,0.85,0.95}
\usepackage{amsmath}
\usepackage{amssymb}
\DeclareMathOperator*{\argmax}{argmax}
\DeclareMathOperator*{\argmin}{argmin}
\usepackage{diagbox}

\usepackage{algpseudocode}
\usepackage{algorithm}
\begin{document}
\title{Robustness-via-Synthesis: Robust Training with Generative Adversarial Perturbations}


\author{\.{I}nci~M.~Bayta\c{s},
        and Debayan~Deb
        % <-this % stops a space
\IEEEcompsocitemizethanks{\IEEEcompsocthanksitem \.{I}nci~M.~Bayta\c{s} is with the Department
of Computer Engineering, Bo\c{g}azi\c{c}i University, \.{I}stanbul, Turkey.
\protect\\
E-mail: inci.baytas@boun.edu.tr
\IEEEcompsocthanksitem Debayan Deb is with LENS Inc., Okemos, MI, USA.
\protect\\
E-mail: debayan@lenscorp.ai}% <-this % stops an unwanted space
\thanks{}}


\markboth{}%
{Shell \MakeLowercase{\textit{et al.}}: Robustness-via-Synthesis: Robust Training with Generative Adversarial Perturbations}




\maketitle

\begin{abstract}
Upon the discovery of adversarial attacks, robust models have become obligatory for deep learning-based systems. Adversarial training with first-order attacks has been one of the most effective defenses against adversarial perturbations to this day. The majority of the adversarial training approaches focus on iteratively perturbing each pixel with the gradient of the loss function with respect to the input image. However, the adversarial training with gradient-based attacks lacks diversity and does not generalize well to natural images and various attacks. This study presents a robust training algorithm where the adversarial perturbations are automatically synthesized from a random vector using a generator network. The classifier is trained with cross-entropy loss regularized with the optimal transport distance between the representations of the natural and synthesized adversarial samples. Unlike prevailing generative defenses, the proposed one-step attack generation framework synthesizes diverse perturbations without utilizing gradient of the classifier's loss. Experimental results show that the proposed approach attains comparable robustness with various gradient-based and generative robust training techniques on CIFAR10, CIFAR100, and SVHN datasets. In addition, compared to the baselines, the proposed robust training framework generalizes well to the natural samples. Code and trained models will be made publicly available.
\end{abstract}


% \fi
\IEEEpeerreviewmaketitle

\section{Introduction}\label{sec:intro}
%!TEX root = main.tex
\IEEEPARstart{D}{eep} neural networks, in particular, convolutional neural networks (CNNs) have become a cornerstone for representation learning in various computer vision applications. Remarkable state-of-the-art performances of CNN architectures have been reported for various benchmark datasets in literature~\cite{alzubaidi2021review}. On the other hand, Szegedy {\it et al.}~\cite{szegedy2013intriguing} unraveled an unprecedented vulnerability of CNNs to specifically crafted, but imperceptible, perturbations known as adversarial attacks. Upon this discovery, various studies~\cite{papernot, carlini2017towards,wiyatno2018maximal,moosavi2015deepfool,advface} showed that it is possible to generate adversarial perturbations of varying strengths using a target, or a surrogate deep model. 

The adversarial attacks can be classified into three categories, such as white-box, gray-box, and black-box attacks~\cite{attacks}. The white-box attacks require the full knowledge of the target model~\cite{attacks}. Projected Gradient Descent (PGD)~\cite{madry2017towards} is accepted as one of the most powerful white-box attacks. The PGD method iteratively computes the first-order gradient of the target model with respect to the input. Whereas, the gray-box attacks have access to the architecture of the target model however the learned model parameters are not available~\cite{attacks}. On the other hand, the black-box methods are only allowed to make queries to the model~\cite{attacks}. The presence of various types of adversarial attacks indicates that the adversarial samples may emerge as a security risk for many deep learning-based systems, such as self-driving cars~\cite{car}, face recognition systems~\cite{face}, and healthcare~\cite{finlayson2019adversarial}. Therefore, it is now imperative to safeguard deep networks against adversarial perturbations. 

In a standard supervised training setting, CNNs are trained with a set of natural images which lack representation of potential adversarial samples. Thus, a naturally trained deep classifier is prone to misclassify the adversarial samples. To alleviate the vulnerability of the CNNs against adversarial samples, adversarial defense methods have been developed. The state-of-the-art defense techniques pose the adversarial robustness as a generalization problem and try to regularize the deep model training by augmenting the training set with adversarial samples. Such techniques are known as {\it adversarial training}~\cite{goodfellow2014explaining, madry2017towards}. In the most common adversarial training methods~\cite{goodfellow2014explaining, madry2017towards}, the adversarial samples are generated by a single type of adversarial attack during the training. 

Although adversarial training is accepted as one of the most effective adversarial defense techniques that is applied to various domains~\cite{graphAT,multiAT}, it has received criticism due to several issues. First, the adversarial training conditions the robustness on only one type of attack~\cite{Tramer2019}. Unfortunately, both single step and iterative adversarial generation techniques fail to find transferable attacks~\cite{zhou2018transferable}. Moreover, adversarial training with a strong iterative attack does not necessarily correspond to improved robustness against various kinds of attacks~\cite{zhang2019theoretically}. Therefore, adversarial training suffers from overfitting certain first-order adversaries. Second, the adversarial training with strong attacks forces the model to capture certain features on extremely perturbed images. As a result, the test performance on natural images degrades severely as the attack strength in adversarial training increases. Lastly, the adversarial training techniques with first-order attacks are computationally expensive since extra back-propagation steps are essential to generate adversarial samples at each training step. Therefore, the quest for a more generalizable adversarial defense technique with less computational overhead and pristine natural accuracy abides.

To alleviate the overfitting issue, we propose synthesizing diverse adversarial perturbations during adversarial training. Generative models can be utilized to introduce diversity in the adversarial perturbations. Although generative methods are extremely versatile in modeling distributions, robust models trained with generative attacks~\cite{jiang2021learning,jang2019adversarial,jeddi2020learn2perturb} has not reached the desired level of adversarial robustness and generalizability to natural samples compared to adversarial training with gradient-based attacks. 

Instead, we propose a generative method that synthesizes one-step diverse perturbations without utilizing gradient of the classifier's loss with respect to input image. The proposed generator is incorporated into a regularized adversarial end-to-end training framework for enhanced robustness to perturbations while maintaining superior generalizability to natural samples.  

The contributions of the study are summarized below:
\begin{itemize}
    \item A lightweight generator is designed to synthesize adversarial perturbations from a random vector. Without utilizing the input image, the generator can output diverse adversarial perturbations.
    \item The generator is trained to maximize the optimal transport distance between natural and synthesized adversarial samples. Therefore, the perturbation generation process does not depend on the class labels.
    \item The perturbation generator does not require the gradient of the classifier's loss function with respect to the input image. In addition, the generator synthesizes adversarial attacks in a single step. Thus, the proposed robust training has less time complexity than adversarial training with iterative attacks.
    \item The objective of the proposed robust classifier is regularized with optimal transport distance between the natural and synthesized adversarial samples. Thus, the representation learning layers are guided to output features that follow similar distributions for natural images and their adversarial counterparts. 
    \item The proposed end-to-end robust training approach does not sacrifice the natural accuracy while providing a robustness that is comparable with the state-of-the-art adversarial accuracy in literature.
\end{itemize}
The rest of this paper is organized as follows. In Section~\ref{sect:related}, we overview state-of-the-art adversarial training methods with gradient-based and generative attacks. In Section~\ref{sect:method}, the proposed framework is presented. The experimental results and analysis are discussed in Section~\ref{sect:exp}. Final discussion is provided in Section~\ref{sect:conc}.

\begin{figure*}[!t]
    \centering
    \includegraphics[width=1.0\textwidth]{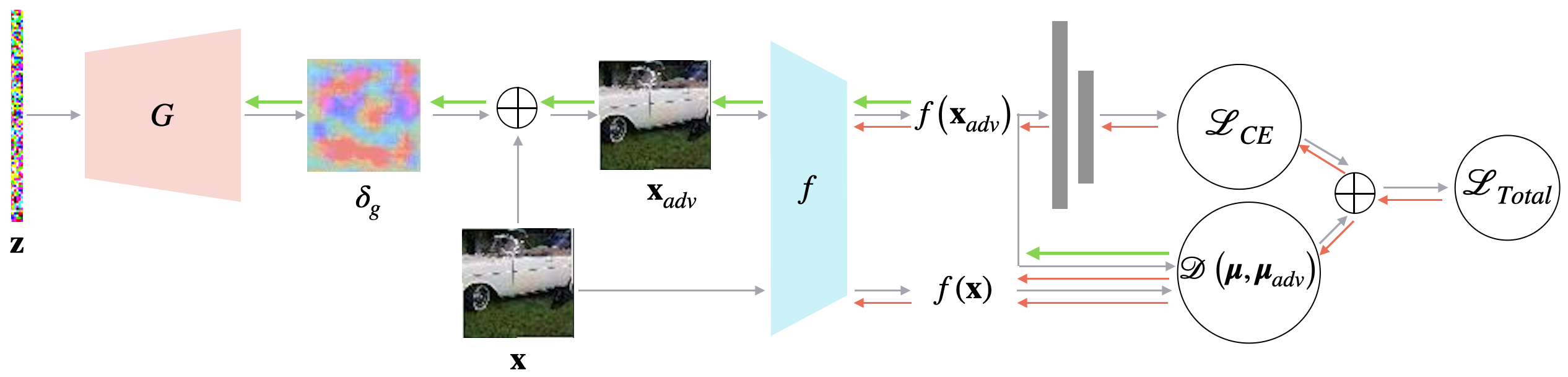}
    \caption{Overview of the robustness via synthesis framework. Input of the generator is a normal random vector denoted by $\mathbf{z}$. The perturbation obtained by the generator $G$ is denoted by $\delta_{g}$. The adversarial sample is computed by adding $\delta_{g}$ to the original image $\mathbf{x}$. The latent representations of the original $f\left(\mathbf{x}\right)$ and the adversarial images $f\left(\mathbf{x}_{\text{adv}}\right)$ are obtained using a CNN architecture. The red and green arrows represent the path of error propagation to update the weights of $f$ and $G$, respectively. The generator is updated using the optimal transport distance $\mathcal{D}\left(\boldsymbol\mu, \boldsymbol \mu_{\text{adv}}\right)$ between the latent representations of the original and the adversarial images. On the other hand, the classifier CNN is updated using the cross-entropy loss $\mathcal{L}_{CE}$ regularized by $\mathcal{D}\left(\boldsymbol\mu, \boldsymbol \mu_{\text{adv}}\right)$.}
    \label{fig:model}
\end{figure*}

\section{Related Work}
\label{sect:related}
%!TEX root = main.tex
The PGD attack has been one of the most effective white-box first-order attacks in literature~\cite{madry2017towards}. As a result, PGD adversarial training is one of the most effective adversarial defense techniques. On the other hand, the PGD adversarial training has a poor generalization performance due to the lack of diversity in the PGD attacks~\cite{yuksel2020adversarial}. There is a growing number of studies that investigate more robust and more generalizable adversarial defense techniques. While some of the studies aim to generate more diverse gradient-based attacks~\cite{zhang2019defense, lee2020adversarial, zhang2020adversarial}, some of them obtain adversarial samples through generative processes~\cite{jang2019adversarial, jeddi2020learn2perturb, jiang2021learning}. In this section, we overview studies that focus on improving the robustness and generalizability of adversarial training with gradient-based and generative attacks.

\subsection{Adversarial training with gradient-based attacks}
Various studies in the literature have analyzed the trade-off between adversarial robustness and generalization. For instance, Zhang and Wang~\cite{zhang2019defense} discussed that the common issues in adversarial training, such as label leaking, are due to solely focusing on the decision boundary during attack generation. To facilitate a more generalizable adversarial training, adversarial perturbations are obtained considering inter-sample relationships rather than the decision boundary~\cite{zhang2019defense}. This feature scattering-based method generates adversarial samples via maximizing the difference between the distributions of natural and adversarial samples over a mini-batch. The feature scattering-based attack~\cite{zhang2019defense} enables improved robustness with a single-step gradient-based attack compared with the standard PGD adversarial training~\cite{madry2017towards}. Wang and Zhang~\cite{wang2019bilateral} also proposed bilateral adversarial training where both image and label are perturbed during training. The authors adopted targeted attack with the most confusing class.

Some studies argue that the existence of the adversarial samples is due to the features with non-robust components. Lee~\cite{lee2020adversarial} proposed a vertex mixup approach to alleviate overfitting to the non-robust features. Their proposed adversarial vertex mixup approach comprises label smoothing and data augmentation. Label smoothing alleviates overfitting by making the model less confident about its predictions. Meanwhile, the training set is diversified through the combinations of adversarial and original samples, where the adversarial samples are obtained with PGD. Another interpolation technique to improve the adversarial robustness was studied by Zhang and Xu~\cite{zhang2020adversarial}. In their study, adversarial interpolation~\cite{zhang2020adversarial} is computed by minimizing the distance between the feature representations of a randomly perturbed sample and an adversarial sample, which is generated by maximizing the cross-entropy loss function. The interpolated input and target pairs are used in adversarial training. Adversarial interpolation is later used along with generative attacks to boost the generalization of the robust training~\cite{jiang2021learning}.

Empirical analysis demonstrates that the generalization property of adversarial training can be improved without sacrificing robustness with methods such as Feature Scatter~\cite{zhang2019defense} and Adversarial Interpolation Training~\cite{zhang2020adversarial}. However, the challenges due to the complexity of incorporating a gradient-based attack and overfitting to a single type of attack remain. For this reason, the quest for more diverse and efficient attacks to integrate into the robust training leads the way to the generative models.

\subsection{Adversarial training with generative attacks}
White-box gradient-based attacks have gained recognition as the most powerful adversarial perturbation generation technique. On the other hand, incorporating powerful adversarial attacks in robust training may not always reflect strong robustness. Since multi-step gradient attacks are prone to lack diversity, generative models have been considered an option for exploring adversarial samples in the input space.

Generative Adversarial Trainer (GAT) by Lee {\it et al.}~\cite{lee2017generative} is one of the early defense techniques that replace the sign of the input gradient with a generator, which is conditioned on the input gradient. The amount of the perturbation is controlled by the $\ell_2$ norm regularization. After the generator is updated, the training set is augmented with the adversarial samples. The discriminator's objective function is the same as the objective function with adversarial regularization proposed by Goodfellow {\it et al.}~\cite{goodfellow2014explaining}. Samangouei {\it et al.}~\cite{samangouei2018defense}, on the other hand, posed the robustness against adversarial attacks as denoising adversarial samples encountered during inference. Therefore, the authors designed Defense-GAN, which can be used with any classifier. At test time, the Defense-GAN generates a reconstruction of the original input without the adversarial noise. However, the empirical evidence for the denoising capabilities of the Defense-GAN is only demonstrated for one-step FGSM and CW attacks. For this reason, the efficiency of the Defense-GAN against multi-step attacks is unclear.

One of the factors that cause poor generalization is using one type of attack to solve the inner maximization problem of the adversarial training. Dong {\it et al.}~\cite{NEURIPS2020_5de8a360} addressed this issue by learning an adversarial distribution rather than a single adversarial sample to solve the inner maximization problem. Their proposed solution, namely Adversarial Distribution Training (ADT), models the distribution of the potential adversarial samples around each input. Compared with the state-of-the-art results on benchmark datasets, the performance of the ADT is inferior. However, the level of robustness provided by ADT is more consistent across various attacks.

In a more recent study, Jeddi {\it et al.}~\cite{jeddi2020learn2perturb} proposed a robust training framework, namely Learn2Perturb, by injecting perturbations into each layer to increase the uncertainty of the deep network. The perturbations, which are added to each feature map, are drawn from a zero-mean normal distribution with a learnable standard deviation. Learn2Perturb alternatingly updates the parameters of the model and perturbation-injection. Unlike the traditional generation of adversarial perturbations, the Learn2Perturb aims to learn the distribution of the perturbations in the latent spaces governed by the feature maps. On the other hand, the Learning2Perturb model cannot overcome the poor generalization performance of the PGD adversarial training considering its test performance on natural images. 

Jiang {\it et al.}~\cite{jiang2021learning} on the other hand, adopted a generic learning-to-learn (L2L) framework for adversarial training. The authors designed an attacker network to generate perturbations. The attacker network either synthesizes perturbations via only the original samples or concatenation with their gradients. The latter is called gradient attacker, and its multi-step version is also provided~\cite{jiang2021learning}. The multi-step gradient attacker aims to mimic the PGD attack with an RNN. Although one-step and two-step gradient attackers improve the adversarial accuracy, they still cannot generalize well to the natural samples. However, Jiang {\it et al.}~\cite{jiang2021learning} also showed that combining L2L with adversarial interpolation training yields a boost in both natural and adversarial accuracies.

L2L based robust training is also investigated by Jang {\it et al.}~\cite{jang2019adversarial}. The authors proposed a generator that can synthesize strong and diverse attacks. Their proposed framework, named L2L-DA~\cite{jang2019adversarial}, adopts a recursive approach to generate strong perturbations while the diversity is enforced by an additional diversity loss. Unlike the gradient attacker in L2L~\cite{jiang2021learning}, the generator takes random noise in addition to the original sample and its gradient. Experiments on benchmark datasets indicate improved robustness compared to L2L~\cite{jiang2021learning}. However, both robust and natural accuracies of L2L-DA cannot reach state-of-the-art performances.

Wang and Yu~\cite{wang2019direct} also designed a GAN to parametrize the inner maximization problem. A more traditional GAN structure is considered to generate adversarial perturbations. Similar to L2L-DA~\cite{jang2019adversarial} and L2L~\cite{jiang2021learning}, the input of the generator is the original sample. The $tanh$ activation function in the last layer of the generator ensures that the perturbations will not exceed the epsilon ball. The author also suggested regularizing the discriminator network with the norm of the gradient in order to stabilize the GAN training. Their proposed robust training cannot improve the adversarial accuracy over PGD adversarial training, however, it achieves a much higher natural accuracy than the PGD adversarial training.

The majority of the studies reviewed in this section require the gradient of the classifier's loss. Even the generative perturbation frameworks take gradient information along with the input sample and utilize recurrent architectures to intensify the adversarial perturbations. However, to the best of our knowledge, robust training techniques with generative frameworks sacrifice either robustness or generalization to natural samples in order to increase the attack strength and diversity. In this study, we present an adversarial perturbation generation method that promotes diversity without utilizing any input sample or gradient information. Yet, the proposed framework provides a more robust and generalizable classifier than the adversarial training techniques with generative attacks discussed in this section. In this regard, we posit that regularizing the classifier is crucial to lead the generator to explore diverse adversarial directions. Updating the classifier with  cross-entropy loss alone is not sufficient to learn a robust representation even if we augment the training set with strong generative attacks.

\section{Method}
\label{sect:method}
%!TEX root = main.tex

% Generalization to natural samples property of a model is essential for learning from data. Unfortunately, we do not have an access to all possible samples that lie in the input space, including the adversarial ones. 
Due to the fact that we do not have access to all possible samples that lie in the input space (including adversarial examples), the standard training procedure inevitably encodes non-robust features. Therefore, if we require any robustness against adversarial directions in the input space, it is imperative to introduce diverse perturbations to the classifier during training. To promote diversity in perturbations, we propose a regularized adversarial training technique that encourages the proposed generator to synthesize diverse attacks.

\subsection{Problem Definition}

Given a dataset $\mathcal{D}=\{\left(\mathbf{x}_i, y_i\right)\}_{i=1}^{N}$ of $N$ natural images $\mathbf{x}_i \in \mathbb{R}^{d_w\times d_h \times c_{in}}$ and their labels $y_i \in {1, \cdots, C}$, the adversarial training is posed as the following two step optimization problem:
\begin{align}
\min_{\boldsymbol{\theta}} \frac{1}{N}\sum_{i=1}^{N} \max_{\boldsymbol{\delta}_i \in S} \mathcal{L} \left(f\left(\mathbf{x}_i + \boldsymbol{\delta}_i; \boldsymbol{\theta}\right), y_i\right)
    \label{eq:adv_training}
\end{align}
where $\boldsymbol{\theta}$ is the parameter of the classifier network, $\boldsymbol{\delta}_i$ is the perturbation generated for $\mathbf{x}_i$, $S=\{\boldsymbol{\delta}: ||\boldsymbol\delta||_{\infty}\leq \epsilon\}$ is the set of allowed perturbations, and $\mathcal{L}\left(\cdot\right)$ is an objective function, e.g, cross-entropy. 

The perturbation $\boldsymbol \delta$ is generated via the inner maximization step in the Eq.~\ref{eq:adv_training}, which is an intractable problem. The gradient of the objective function at an input data point is extremely informative about the adversarial direction. For this reason, one of the most powerful adversarial attack approaches is Projected Gradient Descent (PGD)~\cite{madry2017towards} given below.
\begin{align}
\mathbf{x}^{t+1} = \Pi_{\mathbf{x}+S}\left(\mathbf{x}^t + \alpha \text{sign}\left(\nabla_{\mathbf{x}}\mathcal{L} \left(f\left(\mathbf{x}_i + \boldsymbol{\delta}_i; \boldsymbol{\theta}\right), y_i\right)\right)\right)
\label{eq:pgd}
\end{align}
where $\Pi$ is a projection ensuring that the amount of perturbation will not exceed $\epsilon$ and the adversarial sample will be inside the input domain, $\alpha$ is the step size, and $\text{sign}\left(\nabla_{\mathbf{x}}\mathcal{L} \left(f\left(\mathbf{x}_i + \boldsymbol{\delta}_i; \boldsymbol{\theta}\right), y_i\right)\right)$ is the direction at the input that increases the value of the objective function. Thus, the PGD adversarial training~\cite{madry2017towards} poses robust training as a saddle point problem, where the PGD attacks are obtained by the maximization step and the model parameters are updated via the minimization problem. 

There are several issues with adversarial training when the attack method is PGD. Although the PGD attack is quite strong, the diversity among the attacks at different training iterations is limited. Thus, label leaking and catastrophic overfitting often hinder both the robust and generalization performance of the PGD adversarial training. Besides, when the perturbations obtained by PGD are added to the input image, we may lose the information of the natural patterns. We can observe this issue by the degradation in the natural accuracy of the adversarially trained models. Furthermore, iterative gradient-based attacks are very time-consuming due to multiple backpropagation steps at each iteration. 

In this study, the proposed adversarial training framework, illustrated in Fig.~\ref{fig:model}, is designed to induce adversarial robustness without sacrificing the natural accuracy. In particular, the proposed approach aims to alleviate the following challenges.
\begin{itemize}
    \item Although the iterative gradient-based adversarial attacks~\cite{madry2017towards} are strong regarding the reduction in test accuracy, they often push the natural samples towards similar adversarial directions such that the perturbations may be constantly projected back onto the boundary of $\ell_{\infty}$ norm ball. Thus, the adversarial perturbations are not versatile enough to enable robustness without overfitting to a certain type of attack.
\item Iterative gradient-based adversarial attack generation algorithms are computationally expensive. Therefore, they may not scale to large-scale problems. 
    \item In robust training with generative attacks, when the attack generator network is conditioned on a single input or its gradient, the generator network may overfit to a certain subtle perturbation around the input sample. This situation gets more acute when the parameters of the generator are updated by maximizing the cross-entropy loss.
\item Updating both attack generator and the classifier networks with cross-entropy loss~\emph{alone} results in weaker perturbations.  
\end{itemize}
The proposed framework comprises two modules: adversarial perturbation generator and a classifier. In the next section, adversarial perturbation generator will be discussed.

\subsection{Adversarial Perturbation Generator}
Generative Adversarial Networks (GANs) have been one of the most popular generative models that indirectly learn the distribution of the input data. The generator network may take a random vector or may be conditioned on a specific sample and generates realistic fake samples. A well-trained binary classifier that discriminates a sample as fake or real, is necessary to induce the generator to output more realistic samples~\cite{goodfellow2014generative}. Inspired by the GAN mechanism, we propose an adversarial perturbation generator $G\left(\mathbf{z}; \boldsymbol{\Phi}\right)$ where $\mathbf{z}$ is a random vector and $\boldsymbol{\Phi}$ is the parameters of the generator.

The main purpose of using a generator to obtain adversarial perturbations is to learn how the perturbations within an $\epsilon$ ball are distributed. As seen in Fig.~\ref{fig:model}, unlike many generative attack models in literature, the proposed generator $G\left(\mathbf{z}; \boldsymbol{\Phi}\right)$ is not conditioned on the input sample or the gradient of the loss function at the input. There are two essential reasons behind this design. First, the proposed generator does not output an adversarial image but an adversarial perturbation. For this reason, the output of the generator is not explicitly forced to be visually similar to a particular pattern. Consequently, generating the perturbation from a random vector facilitates diversity among adversarial samples compared with generating a perturbation at a particular data point. Second, decoding a random vector into a perturbation tensor is less computationally expensive than including an extra encoder in a pixel to pixel setting. 

To generate an adversarial attack, we first sample a normal random vector, $\mathbf{z} \in \mathbb{R}^d$, in the latent space of the classifier. The random vector is then decoded into a tensor of the same size as the input image $\mathbf{x}$ as shown in Table~\ref{tab:generator}. This tensor has unbounded values. However, we avoid using the $\tanh$ activation function in the last layer of the generator. Due to its flat regions, the $\tanh$ function is notoriously prone to numerical instabilities. Moreover, according to our observations, the $\tanh$ activation function forces the majority of the perturbation values to be exactly $-1$ and $1$ that prevents the diversity. To ensure that the perturbation is within the set of allowed perturbations in the $\ell_{\infty}$ ball, clipping between the values $\left[-\epsilon, \epsilon\right]$ is employed instead of multiplying the $\tanh$ output by $\epsilon$. Thus, the generator is intended to be more flexible to explore potential adversaries in a wider territory. 

\begin{table}[!t]
\caption{Generator architecture. The latent representation before the last fully connected layer of the WRN-28-10 network is $640$ dimensional. The CIFAR10, CIFAR100, and SVHN datasets have RGB images of size $32\times 32\times 3$. For this reason, a normal random vector is drawn from the $640$ dimensional space and decoded to a $32\times32\times3$ tensor.}
    \label{tab:generator}
    \centering
    \begin{tabular}{|c|c|c|}
    \hline
        \textbf{Layer} & \textbf{Size} & \textbf{Output}\\
        \hline
        \hline
        Input & $\mathbf{z}\in \mathbb{R}^{640} \sim \mathcal{N}\left(0,1\right)$ & $640 \times 1$\\
        \hline
        FC & $640 \times 4096$ & $4096 \times 1$ \\
        \hline
        Reshape & & $8\times 8\times 64$ \\
        \hline
        DeConv & Kernel:$4\times4$, Stride:$2$, Channels:$32$ & $16\times 16\times 32$ \\
        \hline
        Batchnorm & & \\
        \hline
        Leaky ReLU & Leakiness:$0.2$ & \\
        \hline
        DeConv & Kernel:$4\times4$, Stride:$2$, Channels:$16$ & $32\times 32\times 16$ \\
        \hline
        Batchnorm & & \\
        \hline
        Leaky ReLU & Leakiness:$0.2$ & \\
        \hline
        Conv & Kernel:$4\times4$, Stride:$1$, Channels:$3$ & $32\times 32\times 3$ \\
        \hline
    \end{tabular}
\end{table}

After obtaining the perturbation $\boldsymbol\delta $, the adversarial sample is obtained as follows 
\begin{align}
\boldsymbol\delta_{g} &= \Pi_{S}\left(G\left(\mathbf{z};\boldsymbol\Phi\right)\right) \\
\mathbf{x}_{\text{adv}} &= \Pi_{\mathbf{x}}\left(\mathbf{x} + \boldsymbol\delta_{g}\right)
    \label{eq:adv_sample}
\end{align}
where $\mathbf{x}_{\text{adv}} \in \mathbb{R}^{d_w \times d_h \times c_{in}}$ is the adversarial sample, $\Pi_{S}$ and $\Pi_{\mathbf{x}}$ denote the projection operators (e.g., clipping) to map the perturbation into $S$ and the adversarial sample back into the input domain, respectively.

Parameters of the generator, $\boldsymbol\Phi$ are updated through the following optimization problem.
\begin{align}
\max_{\boldsymbol\Phi} \mathcal{D}\left(\boldsymbol\mu, \boldsymbol\mu_{\text{adv}}\right)  
    \label{eq:gan_update}
\end{align}
where $\mathcal{D}\left(\boldsymbol\mu, \boldsymbol\mu_{\text{adv}}\right)$ denotes the optimal transport (OT) distance between the natural and the adversarial sample distributions. The OT distance is a well-studied distance that stabilizes and improves the GAN training~\cite{genevay2017gan, salimans2018improving}. The OT distance between two distributions is defined as follows.
\begin{align}
\mathcal{D}\left(\boldsymbol\mu,\boldsymbol\nu\right) = \inf_{\gamma \in \Pi\left(\boldsymbol\mu, \boldsymbol\nu\right)} E_{\left(\mathbf{x}, \mathbf{y}\right) \sim \gamma} c\left(\mathbf{x}, \mathbf{y}\right) 
    \label{eq:ot}
\end{align}
where $\Pi\left(\boldsymbol\mu, \boldsymbol\nu\right)$ is the set of joint distributions $\gamma\left(\mathbf{x},\mathbf{y}\right)$ with marginal distributions of $\boldsymbol\mu\left(\mathbf{x}\right)$ and $\boldsymbol\nu\left(\mathbf{y}\right)$, and $c\left(\mathbf{x},\mathbf{y}\right)$ is a cost function~\cite{zhang2019defense}. The OT distance represents the minimum cost to transport one marginal to another. 

Following the footsteps of Zhang {\it et al.}~\cite{zhang2019defense}, finding the OT distance is equivalent to the following problem.
\begin{align}
\mathcal{D}\left(\boldsymbol\mu, \boldsymbol\mu_{\text{adv}}\right)  &= \nonumber \\ 
\min_{\mathbf{T}\in \Pi\left(\mathbf{u}, \mathbf{u}_{\text{adv}}\right)} &\sum_{i=1}^{n}\sum_{j=1}^{n}\mathbf{T}_{ij}\cdot c\left(f\left(\mathbf{x}\right)_i, f\left(\mathbf{x}_{\text{adv}}\right)_j\right) 
\label{eq:ot_reg}
\end{align}
where $n$ denotes the size of the mini-batch, $\Pi\left(\mathbf{u}, \mathbf{u}_{\text{adv}}\right)=\{\mathbf{T}\in \mathcal{R}_{+}^{n\times n}| \mathbf{T}\mathbf{1}_n = \mathbf{u}, \mathbf{T}^{T}\mathbf{1}_{n}=\mathbf{u}_{\text{adv}}\}$, $\mathbf{1}_n$ is $n-$dimensional all-one vector, $\mathbf{u}$ and $\mathbf{u}_{\text{adv}}$ contain the values of the weight vectors $\boldsymbol\mu=\{u_i\}_{i=1}^{n}$ and $\boldsymbol\mu_{\text{adv}}=\{v_i\}_{i=1}^{n}$, respectively \cite{zhang2019defense}. Since $\boldsymbol\mu$ and $\boldsymbol\mu_{\text{adv}}$ are probability distributions, $\sum_{i}^{n} u_i = \sum_{i=1}^{n}v_i=1$. In this study, the cost function in Eq.~\ref{eq:ot_reg} is defined as the euclidean distance between the latent representations of the natural and the adversarial samples as shown below.
\begin{align}
c\left(f\left(\mathbf{x}\right)_i, f\left(\mathbf{x}_{\text{adv}}\right)_j\right)= \lVert f\left(\mathbf{x}\right)_i - f\left(\mathbf{x}_{\text{adv}}\right)_j \rVert_{2}^{2}
\label{eq:cost}
\end{align}
 
The OT distance in Eq.~\ref{eq:ot_reg} is coupled over mini-batches such that the distance between empirical distributions of the natural samples and their adversarial counterparts in one mini-batch is measured. Since the adversarial perturbation is not generated at a single sample, the generation process potentially explores a wider region in the input space. Secondly, the perturbation is not generated by taking the sign of the gradient of the OT distance. When we add the sign of the gradient to the input sample, every pixel moves exactly the same amount. For the pixel values that are close to the upper and lower bounds of the input domain, the perturbation may not have any effect due to clipping. Such cases are prone to overfitting. Therefore, a generator is designed to output a perturbation tensor whose values range between $\left[-\epsilon, \epsilon\right]$.

The proposed generator does not require multiple backpropagation steps or a recurrent loop. Thus, the proposed adversarial perturbation generator is less expensive than iterative attacks. As presented in Algorithm~\ref{alg:pgd}, at each iteration, the perturbation generator is updated once. The generator weights are updated in order to maximize the OT distance between the latent representations of natural and synthesized adversarial samples. As the classifier becomes more robust during the training, the perturbation generator will be encouraged to explore more sophisticated perturbations. In the next section, details of the classifier training are presented. 

\subsection{Classifier}
In a standard GAN setting, a discriminator is trained to classify fake and real images while providing informative gradients to the generator. In this study, unlike the standard-setting, a classifier is trained to accomplish an object recognition task. 
% On the other hand, similar to the standard setting, the error backpropagating through the classifier is used in the generator update. 
Inspired by the GAN pipeline, the OT distance loss is backpropagated through the classifier to update the  generator's parameters. For this reason, the status of the classifier during the training enables the generator to create diverse adversarial perturbations. 
% We find that utilizing the classifier's cross-entropy loss for updating the generator leads to the generator overfitting. 
Furthermore, the goal of robust training is to learn a latent space in which the feature representations of the natural sample and its adversarial counterpart do not differ. However, updating the classifier's parameters based solely on the cross-entropy loss may not achieve this objective. 

To alleviate the challenge discussed above, the OT distance between the natural and the synthesized adversarial sample distributions is used as a regularizer. The total loss function of the classifier is given below.
\begin{align}
\mathcal{L}_{\text{total}}\left(\mathbf{x}, \mathbf{x}_{\text{adv}}, y; \boldsymbol\theta, \boldsymbol\Phi\right)=\mathcal{L}_{CE}\left(\mathbf{x}_{\text{adv}}, y;\boldsymbol\theta\right) + \mathcal{D}\left(\boldsymbol\mu, \boldsymbol\mu_{\text{adv}}\right)
    \label{eq:total_loss}
\end{align}
where $\mathcal{L}_{CE}\left(\mathbf{x},y; \boldsymbol{\theta}\right) = -\frac{1}{N}\sum_{i=1}^{N}\sum_{j=1}^{C}y_{ij}\log f_{j}\left(\mathbf{x}_i; \boldsymbol{\theta}\right)$ is the cross-entropy loss function, $\boldsymbol\theta$ denotes the parameters of the classifier, and $\mathcal{D}\left(\boldsymbol\mu, \boldsymbol\mu_{\text{adv}}\right)$ is defined in Eq.~\ref{eq:ot_reg}.

A significant improvement in robustness is observed when the OT distance regularization is included in the loss function of the classifier given in Eq.~\ref{eq:total_loss}. The contributions of the OT distance regularization is two-fold; (i) classifier is encouraged to learn a latent space that is less susceptible to adversarial perturbations, (ii) generator is encouraged to produce more challenging perturbations since the classifier is trained to map adversarial samples near to their natural counterparts in the latent space. 

% In this problem, challenging adversarial perturbations do not necessarily correspond to attack strength in terms of iterative gradient steps. 

In the robust training setting, feeding the classifier with very strong adversarial attacks such as PGD hurts the generalization performance and overfits to the particular attack method. For this reason, adversarial training demands exploring diverse adversarial samples in the input space rather than samples on the boundary of the $\epsilon-$ball that maximizes the objective function. In the next section, the proposed robustness via synthesis framework given in Algorithm~\ref{alg:pgd} is analyzed with experiments on different benchmark datasets.

 \begin{algorithm}[t]
	\caption{Robustness via synthesis for T epochs, M mini-batches, $\boldsymbol\theta$ network parameter, $\boldsymbol\Phi$ generator parameter, and $\mathcal{L}_{\text{total}}$ loss function}
	\begin{algorithmic}[1]
		\For {$ t=1,\cdots,T $}
		\For {$ i=1,\cdots, M $}
		\State Sample a normal random vector, $\mathbf{z}$
		\State $\boldsymbol\delta_g = \Pi_{S}\left(G\left(\mathbf{z};\boldsymbol\Phi\right)\right)$
		\State $\mathbf{x}_{\text{adv}} = \Pi_{\mathbf{x}}\left(\mathbf{x} + \boldsymbol\delta_g\right)$ 
		\State $\boldsymbol\Phi^{*}  = \argmax\limits_{{\boldsymbol\Phi}} \mathcal{D}\left(\boldsymbol\mu, \boldsymbol\mu_{\text{adv}}\right) $
		\State $\boldsymbol\delta_{g} = \Pi_{S}\left(G\left(\mathbf{z};\boldsymbol\Phi^{*}\right)\right)$
		\State $\mathbf{x}_{\text{adv}} = \Pi_{\mathbf{x}}\left(\mathbf{x} + \boldsymbol\delta_{g}\right)$
		\State $\boldsymbol\theta^{*} = \argmin\limits_{\boldsymbol\theta} \mathcal{L}_{\text{total}}\left(\mathbf{x}, \mathbf{x}_{\text{adv}}, y; \boldsymbol\theta, \boldsymbol\Phi^{*}\right)$
		\EndFor
		\EndFor
	\end{algorithmic}
	\label{alg:pgd}
\end{algorithm}

\section{Experiments}
\label{sect:exp}
%!TEX root = main.tex
The robustness of the proposed approach is tested on three common benchmark datasets, namely CIFAR10, CIFAR100~\cite{cifar} and SVHN~\cite{svhn}. In our experiments, object recognition task is considered and classification accuracy is used to compare the proposed and baseline methods. In addition to the white-box and black-box performances, diversity in the perturbation generation and the behavior of the learned latent representations by the proposed model are analyzed in Section~\ref{subsect:diversity}. We also investigate the effect of OT loss regularization and the choice of the loss function to train the generator with an ablation study in Section~\ref{subsect:ablation}.

\begin{table*}[!t]
\caption{White-box Results. In all of the experiments, the maximum amount of perturbation is set to $\epsilon = 8/255$, and the step size is $2/255$. Starred values are obtained by training the corresponding networks from scratch. Other values represent the best performances reported by the baseline studies. The values of the cells with a dash are not available. CIFAR100 results of Learn2Perturb are marked with $\dagger$ to indicate that these values are taken from Figures 5 and 6 in the supplementary material of the study~\cite{jeddi2020learn2perturb}.}
\label{tab:white}
\centering
\begin{tabular}{|c|c|c|c|c|c|c|c|c|c|}
\hline

\multicolumn{10}{|c|}{\rule{0pt}{10pt}\textbf{CIFAR10}}                                                                                                                                                        \\ \hline 
\textbf{Defenses}                                                    & \textbf{Natural} & \textbf{FGSM} & \textbf{PGD-7} & \textbf{PGD-10} & \textbf{PGD-20} & \textbf{PGD-100} & \textbf{CW-20} & \textbf{CW-100} & \textbf{AdvGAN~\cite{xiao2018generating}} \\ \hline 
Natural                                                              & 94.42*            & 10.22*         & 0               & 0               & 0                & 0       &0       & 0 & 10.89*             \\ \hline
AT~\cite{madry2017towards}                     & 87.25            & 62.64   &49.67      & 47.33           & 45.91           & 45.29            & 46.99          & 46.54 & \textbf{84.98*}          \\ \hline
Bilateral~\cite{wang2019bilateral}             & 91.00               & 70.70    & 63.00      & -               & 57.80            & 55.20             & 56.20           & 53.80 & -           \\ \hline
FeatScatter~\cite{zhang2019defense}            & 90.00             & 78.40     &\textbf{73.54*}     & \textbf{70.90}          & 70.50            &  68.60             &  62.40          &  60.60  & -            \\ \hline
Adv-interp~\cite{zhang2020adversarial}         & 90.30             & 78.00   &-       & -               & \textbf{73.50}            & \textbf{73.00}             & \textbf{69.70}           & \textbf{68.70}  & -          \\ \hline
Adv-vertex~\cite{lee2020adversarial}& 93.24             & 78.25   &-       & 62.67               & 58.23            & -             & 53.63           & - & -            \\ \hline
L2L-DA~\cite{jang2019adversarial}              & 78.91            & 45.77   &-      & 39.69           & -               & 38.39            & -               & 37.75    & -       \\ \hline
L2L~\cite{jiang2021learning}                   & 85.35            & -      &-       & -               & 54.32           & 52.12            & -              & 57.07   & -        \\ \hline
Direct~\cite{wang2019direct}                   & 91.08            & 72.81   &-      & 44.28           & -               & -                & -              & -        & -       \\ \hline
Learn2Perturb~\cite{jeddi2020learn2perturb} & 85.30 & 62.43 & 56.06 & - &- &- &-&-&-\\ \hline
\rowcolor{grayblue}\textbf{Ours}                                                        & \textbf{94.17}                 &\textbf{79.99}        & 69.02       &                 65.24 & 57.50                 & 48.62               & 41.21             &27.54  & 72.44               \\ \hline \hline
\multicolumn{10}{|c|}{\rule{0pt}{10pt}\textbf{CIFAR100}}  \\ \hline 
\textbf{Defenses}                                                    & \textbf{Natural} & \textbf{FGSM} &\textbf{PGD-7}& \textbf{PGD-10} & \textbf{PGD-20} & \textbf{PGD-100} & \textbf{CW-20} & \textbf{CW-100} & \textbf{AdvGAN~\cite{xiao2018generating}} \\ \hline 
Natural                                                              & 79.22*            & 3.52*   &0       & 0               & 0               & 0                & 0              & 0  & 4.14*              \\ \hline
AT~\cite{madry2017towards}                     &  59.78*                & 32.70*     &25.07*         &  23.49*               & 22.78*                & 22.44*                 &  23.05*              & 22.87*  & \textbf{55.39*}               \\ \hline
Bilateral~\cite{wang2019bilateral}             & 66.20             & 31.30     &-     & -               & 23.10            & 22.40             & -              & 20.00    & -          \\ \hline
FeatScatter~\cite{zhang2019defense}            & 73.90             & 61.00   &46.29*       & 45.99*               & 47.20            & \textbf{46.20}             & 34.60           & 30.60    & -        \\ \hline
Adv-interp~\cite{zhang2020adversarial}         & 73.6             & 58.3     &-     & -               & 41              & 40.2             & 32.4           & 31.2    & -        \\ \hline
Adv-vertex~\cite{lee2020adversarial}& 74.81             & \textbf{62.76}   &-       & -               & 38.49            & -             & -           & -     & -       \\ \hline
L2L~\cite{jiang2021learning}                   & 60.95            & -    &-         & -               & 31.03           & 29.75            & -              & 32.28   & -        \\ \hline
Direct~\cite{wang2019direct}                   & 70.99            & 41.86  &-       & 18.25           & -               & -                & -              & -       & -        \\ \hline
Learn2Perturb~\cite{jeddi2020learn2perturb}& $58.00^{\dagger}$ & $30.00^{\dagger}$ & $26.00^{\dagger}$ & - &- &- &-&-&- \\ \hline
\rowcolor{grayblue}\textbf{Ours}                                                                 &\textbf{76.32}                  &51.63     &49.50          & \textbf{49.15}                &\textbf{48.16}                 &   45.79               & \textbf{40.60}             &    \textbf{38.25}       & 54.52        \\ \hline \hline
\multicolumn{10}{|c|}{\rule{0pt}{12pt}\textbf{SVHN}}                                                                                                                                                               \\ \hline 
\textbf{Defenses}                                                    & \textbf{Natural} & \textbf{FGSM} &\textbf{PGD-7}& \textbf{PGD-10} & \textbf{PGD-20} & \textbf{PGD-100} & \textbf{CW-20} & \textbf{CW-100} & \textbf{AdvGAN~\cite{xiao2018generating}} \\ \hline 
Natural                                                              & 95.92*            & 13.62*    &0     &      0.18*           & 0               & 0                & 0              & 0   & 45.30*            \\ \hline
AT~\cite{madry2017towards}                     &90.74*                  & 64.50*      &47.87*        & 44.23*                & 41.38*                &   40.37*               &   42.46*             &  41.60*    & 88.10*           \\ \hline
Bilateral~\cite{wang2019bilateral}             & 94.10             & 69.80   &-       & -               & 53.90            & 50.30             & -              & 48.90    & -        \\ \hline
FeatScatter~\cite{zhang2019defense}            & 96.20             & 83.50   &61.88*       & 55.40*               & 62.90            & 52.00             & 61.30           & 50.80   & -         \\ \hline
Adv-interp~\cite{zhang2020adversarial}         & 94.10             & 75.60  &-       & -               & \textbf{65.80}            & \textbf{64.00}               & \textbf{63.40}           & \textbf{60.40}    & -        \\ \hline
Adv-vertex~\cite{lee2020adversarial}& 95.59             & 81.83   &-       & -               &  61.90            & -             & -           & -  & -          \\ \hline
Direct~\cite{wang2019direct} ($\epsilon=0.05$) & \textbf{96.34}            & \textbf{91.51}   &-      & 37.97           & -               & -                & -              & -     & -          \\ \hline
\rowcolor{grayblue}\textbf{Ours}                                                                 &95.50                  & 80.37     &70.17         & \textbf{67.13}                &60.53                 & 53.98                 & 54.21               & 46.33    & 82.67            \\ \hline
\end{tabular}
\end{table*}

\subsection{Datasets and Implementation Details}
We conduct experiments with CIFAR10, CIFAR100~\cite{cifar}, and SVHN~\cite{svhn} benchmark datasets of $32\times 32$ RGB images. The CIFAR10 and SVHN datasets have 10 classes, whereas the CIFAR100 dataset has 100 categories. The CIFAR10 and CIFAR100 datasets have 50k training images and 10k test images. The SVHN dataset has 73,257 training images and 26,032 test images. In our experiments, compatible with the adversarial training literature, WRN-28-10 wide ResNet~\cite{zagoruyko2016wide} classifier is used. The proposed end-to-end robustness-via-synthesis training framework is implemented in Tensorflow within the codebase provided by MadryLab~\cite{cifar10challenge}. Tensorflow implementation of the Sinkhorn algorithm with regularization of 0.01 and 100 iterations~\cite{tfsinkhorn} is used to compute the OT distance. 

For all of the experiments, the maximum perturbation amount is set to $\epsilon=8/255$. For the CIFAR datasets, random cropping and flipping are applied to augment the training sets~\cite{cifar10challenge}. Throughout the experiments, the batch size is fixed to 64. The CIFAR10 and SVHN models are trained for 179k iterations at which the most optimal robustness is observed. The CIFAR100 model is trained for 300k iterations. Since the OT distance is computed over mini-batches, datasets with a large number of categories may require longer training in order to present all the categories to the generator.

 As suggested in literature~\cite{zhang2019defense,wang2019bilateral}, a learning rate scheduling scheme is adopted with transition steps of 60K and 90K iterations. An initial learning rate of 0.1 is used for the CIFAR datasets and 0.01 for the SVHN dataset with a learning rate decay of 0.1. While updating the parameters of the discriminator, label smoothing with a weight factor of 0.5 is applied. The learning rate for the proposed generator architecture, provided in Table~\ref{tab:generator}, is set to 0.01 and 0.0001 for CIFAR datasets, and the SVHN dataset, respectively. A learning rate schedule is not applied during the optimization of the generator. The momentum optimizer is utilized to train both the classifier and the generator.

\subsection{Baselines}
The proposed robustness via synthesis technique is an adversarial training approach with a generative attack. We compare our method to the recent adversarial training techniques that are in the same category as the proposed method, such as L2L-DA~\cite{jang2019adversarial}, L2L~\cite{jiang2021learning}, Learn2Perturb~\cite{jeddi2020learn2perturb}, and Direct~\cite{wang2019direct}. In Table~\ref{tab:white}, the best performances reported by the studies~\cite{jang2019adversarial,jiang2021learning,jeddi2020learn2perturb,wang2019direct} are included. 

Although the proposed framework is designed to improve the performance of adversarial training with generative attacks, we also provide a comparison with the state-of-the-art adversarial training techniques with gradient-based attacks, such as standard adversarial training (AT) proposed by Madry {\it et al.}~\cite{madry2017towards}, TRADES~\cite{zhang2019theoretically}, bilateral adversarial training (Bilateral)~\cite{wang2019bilateral}, feature scatter-based adversarial training (FeatScatter)~\cite{zhang2019defense}, and adversarial interpolation training (Adv-interp)~\cite{zhang2020adversarial}. For CIFAR100 and SVHN, we train the AT~\cite{madry2017towards} model from scratch. In AT~\cite{madry2017towards} experiments, implementation and hyperparameters provided by MadryLab~\cite{cifar10challenge} are utilized. During the adversarial training, PGD-7 attack is utilized. For FeatScatter~\cite{zhang2019defense} experiments, we trained the models from scratch using the Pytorch implementation~\cite{featscat} for CIFAR100 and SVHN to report the PGD-10 results. However, we present the best results reported by the authors when they are available. 

\subsection{Classification Performance}
In this section, we compare the classification accuracy under various types of attacks. In Table~\ref{tab:white}, all the attacks are generated given the architecture and the parameters of the models. In Table~\ref{tab:black}, adversarial attacks are generated using surrogate models and tested on robust models.

\begin{figure*}[!t]
    \centering
    \subfloat[CIFAR10]{\includegraphics[width=0.33\textwidth]{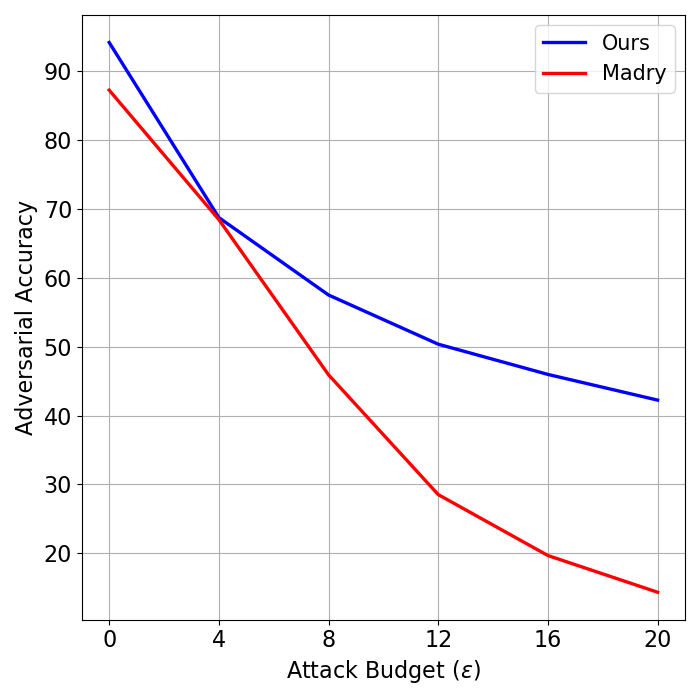}}\hfill
    \subfloat[CIFAR100]{\includegraphics[width=0.33\textwidth]{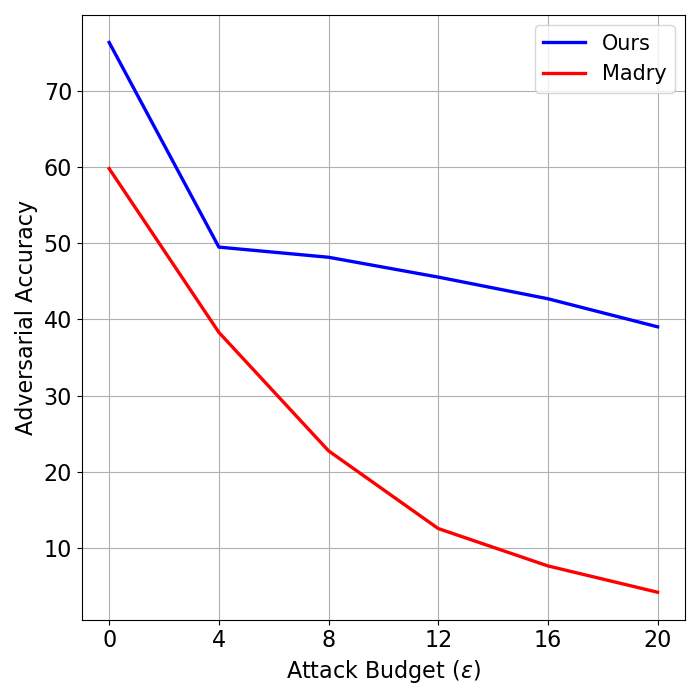}}\hfill
    \subfloat[SVHN]{\includegraphics[width=0.33\textwidth]{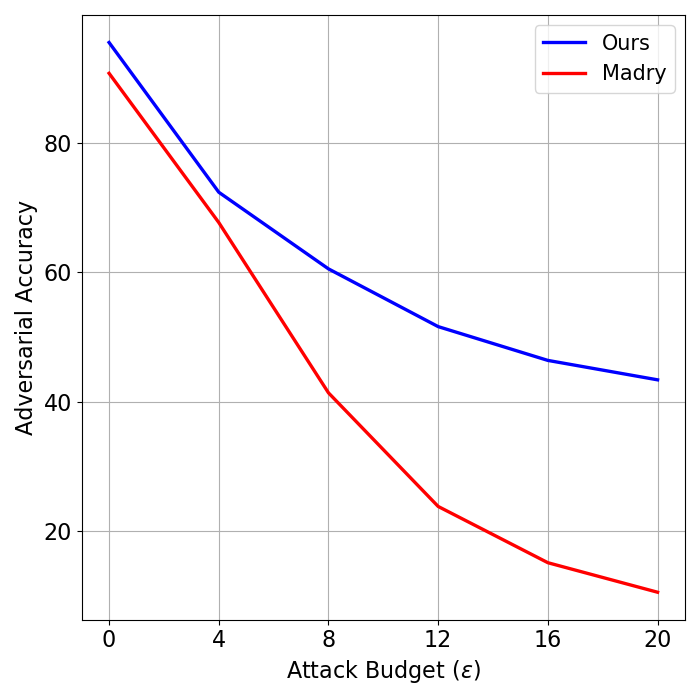}}
    \caption{Robustness across varying attack budgets for (a) CIFAR10, (b) CIFAR100, and (c) SVHN. Adversarial accuracies against the PGD-20 attack of the proposed and Madry's models~\cite{madry2017towards} are plotted for varying $\epsilon \in \left[\frac{0}{255}, \frac{20}{255}\right]$ values. The step size of the PGD attack is set to $\frac{2}{255}$.}
    \label{fig:budget}
\end{figure*}

\subsubsection{White-Box Attack}
PGD~\cite{madry2017towards} is considered as one of the most effective first-order white-box adversarial attacks. Robustness performance of the proposed and the baseline methods are evaluated against FGSM~\cite{goodfellow2014explaining}, PGD-10, PGD-20, and PGD-100 for $\ell_{\infty}$ attacks, and CW-20 and CW-100~\cite{carlini2017towards} for the $\ell_{2}$ norm attacks. The numbers next to the attack types indicate the number of steps. Random initialization is applied to all of the white-box attacks and the step size is set to $2/255$.  

In addition to the gradient-based attacks, white-box performance is also evaluated against a prominent generative attack, namely AdvGAN~\cite{xiao2018generating}. To obtain the adversarial samples via AdvGAN, the generator of the AdvGAN is trained from scratch to fool robust classifiers obtained by the proposed framework, AT~\cite{madry2017towards}, and the natural model. We used the Tensorflow implementation of AdvGAN\footnote{https://github.com/ctargon/AdvGAN-tf}. 
% L2L-DA~\cite{jang2019adversarial} also reports the adversarial accuracy on CIFAR10 against AdvGAN. However, the authors state that the AdvGAN is finetuned to the L2L-DA classifier, and a $75.31 \%$ accuracy is reported. 
In our experiments, we train the generator and the discriminator of the AdvGAN from scratch and report the result at the checkpoint where the adversarial attack is the most successful. 
% In this regard, comparing the performance against AdvGAN reported in the L2L-DA~\cite{jang2019adversarial} paper would not be fair. For this reason, we leave a dash for the L2L-DA result for AdvGAN in Table~\ref{tab:white}. 

As seen in the white-box results, the proposed approach can preserve the natural accuracy across all the datasets. Particularly for CIFAR10 and CIFAR100, generative methods~\cite{jang2019adversarial,jiang2021learning,wang2019direct} fail to generalize to the natural samples and their robustness is poor compared to gradient-based methods. On the other hand, the robustness obtained via the proposed framework is not only better than the other generative approaches and Madry's AT~\cite{madry2017towards}, but also comparable to the state-of-the-art performances by gradient-based techniques such as Feature Scatter~\cite{zhang2019defense}. Reaching this level of robustness and generalizability without the gradient of the cross-entropy loss in the perturbation generation is a notable point. This can be clearly observed when we investigate the change in robustness against PGD-20 for varying attack budgets in Figure~\ref{fig:budget}. We find that the proposed framework can maintain much better robustness than the standard AT even for high $\epsilon$ values. For instance, the proposed framework facilitates preserving the robustness for $\epsilon=20/255$.

\begin{table}[!t]
\centering
\caption{Black-box Results. The Natural Model table presents the robustness of the proposed model against adversarial samples that are generated using the pre-trained natural model. Whereas, Robust Model table contains the performance of the proposed model against adversarial samples generated using the robust model of Madry {\it et al.}~\cite{madry2017towards}.}
\label{tab:black}
\begin{tabular}{cccc}
\multicolumn{4}{c}{\rule{0pt}{12pt}\textbf{Natural Model}}       \\ \hline
\multicolumn{1}{|c|}{\backslashbox{Attacks}{Datasets}}          & \multicolumn{1}{c|}{CIFAR10} & \multicolumn{1}{c|}{CIFAR100} & \multicolumn{1}{c|}{SVHN}   \\ \hline
\multicolumn{1}{|c|}{PGD-20}  & \multicolumn{1}{c|}{71.11} & \multicolumn{1}{c|}{62.10}  & \multicolumn{1}{c|}{82.42}   \\ \hline
\multicolumn{1}{|c|}{PGD-100} & \multicolumn{1}{c|}{71.46} & \multicolumn{1}{c|}{61.51}  & \multicolumn{1}{c|}{83.38}       \\ \hline
\multicolumn{1}{|c|}{CW-20}     & \multicolumn{1}{c|}{71.57} & \multicolumn{1}{c|}{63.57}  & \multicolumn{1}{c|}{82.54}       \\  \hline
\multicolumn{1}{|c|}{CW-100}     & \multicolumn{1}{c|}{71.76} & \multicolumn{1}{c|}{63.77}  & \multicolumn{1}{c|}{82.63}       \\  \hline
\multicolumn{1}{|c|}{AdvGAN~\cite{xiao2018generating}}     & \multicolumn{1}{c|}{85.08} & \multicolumn{1}{c|}{52.19}  & \multicolumn{1}{c|}{95.72}       \\  \hline

\multicolumn{4}{c}{\rule{0pt}{12pt}\textbf{Robust Model}}       \\ \hline
\multicolumn{1}{|c|}{\backslashbox{Attacks}{Datasets}}          & \multicolumn{1}{c|}{CIFAR10} & \multicolumn{1}{c|}{CIFAR100} & \multicolumn{1}{c|}{SVHN}   \\ \hline
\multicolumn{1}{|c|}{PGD-20}  & \multicolumn{1}{c|}{74.53} & \multicolumn{1}{c|}{54.34}  & \multicolumn{1}{c|}{65.69}   \\ \hline
\multicolumn{1}{|c|}{PGD-100} & \multicolumn{1}{c|}{74.12} & \multicolumn{1}{c|}{54.55}  & \multicolumn{1}{c|}{65.16}       \\ \hline
\multicolumn{1}{|c|}{CW-20}     & \multicolumn{1}{c|}{75.31} & \multicolumn{1}{c|}{55.01}  & \multicolumn{1}{c|}{66.69}       \\  \hline
\multicolumn{1}{|c|}{CW-100}     & \multicolumn{1}{c|}{74.96} & \multicolumn{1}{c|}{55.23}  & \multicolumn{1}{c|}{66.46}       \\  \hline
\multicolumn{1}{|c|}{AdvGAN~\cite{xiao2018generating}}     & \multicolumn{1}{c|}{88.22} & \multicolumn{1}{c|}{68.10}  & \multicolumn{1}{c|}{92.58}       \\  \hline
\end{tabular}
\end{table}

\begin{figure}[!t]
    \centering
    \includegraphics[width=0.55\textwidth]{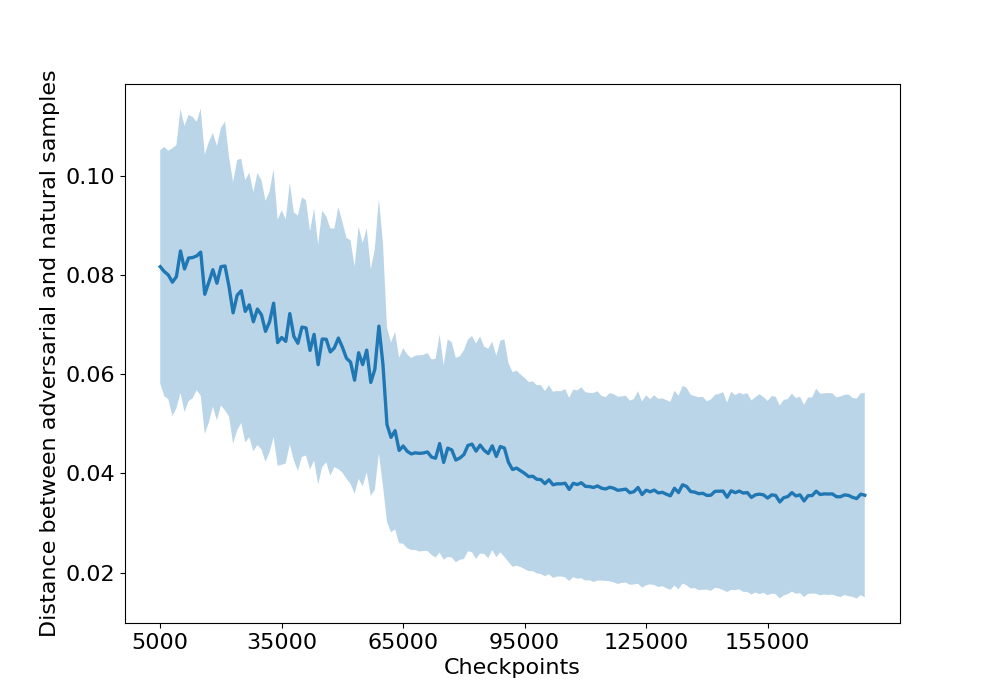}
    \caption{We generate 100 adversarial images for each natural sample in the CIFAR10 test set. The adversarial images are obtained from 100 different random vectors. Then, the Euclidean distances between the normalized latent representations of the natural sample and 100 adversarial samples are computed. The average distances for 174 checkpoints are plotted. The curve represents the change in the average distance during training. The shaded region around the curve represents the average standard deviation of the distances between the natural sample and its 100 adversarial samples. While the diversity in the adversarial directions is preserved during the robust training, the average distance between the latent representations decreases as expected.}
    \label{fig:diverse}
\end{figure}

\subsubsection{Black-Box}
Black-box performance is evaluated against one generative and several gradient-based attacks. The gradient-based black-box attacks are generated using a naturally trained model and robust model that is obtained with the standard adversarial training with PGD-7~\cite{madry2017towards}. Using each model, a set of PGD and CW attacks with step sizes of $20$ and $100$ were generated. Similarly, we generate adversarial samples via AdvGAN using the natural and robust models to evaluate the black-box performance of the proposed model. As seen in Table~\ref{tab:black}, the proposed framework is generalizable to various black-box scenarios.

\subsubsection{Perturbation Diversity Analysis}
\label{subsect:diversity}
To improve the generalizability of the robust model, diverse adversarial samples should be used in the training. One of the motivations for generative perturbations is to explore more diverse adversarial attacks than gradient-based techniques. In this section, we investigate the diversity of the adversarial perturbations generated during the proposed regularized robust training method in the latent space. For this purpose, we generate 100 adversarial images for each natural sample in the CIFAR10 test set. The adversarial images are obtained from 100 different random vectors. Then, the Euclidean distances between the normalized latent representations of the natural sample and 100 adversarial samples are computed. In Figure~\ref{fig:diverse}, we plot the average distances over all the data points in the test set for 174 checkpoints. The curve represents the change in the average distance during training. The shaded region around the curve represents the average standard deviation of the distances between the natural sample and its 100 adversarial samples.

According to our initial hypothesis, there are two expected behaviors in this experiment: (i) The diversity in the adversarial perturbation generation should be preserved during the robust training. (ii) As the robust training progresses, the distances between adversarial samples and the natural samples should decrease since the learned representations are expected to be insusceptible to adversarial perturbations. As it can be observed in Figure~\ref{fig:diverse}, the average distances decrease during the training while the standard deviation of the distances between the natural image and its adversarial samples is preserved.

\begin{figure*}[!t]
    \centering
    \includegraphics[width=1.0\textwidth]{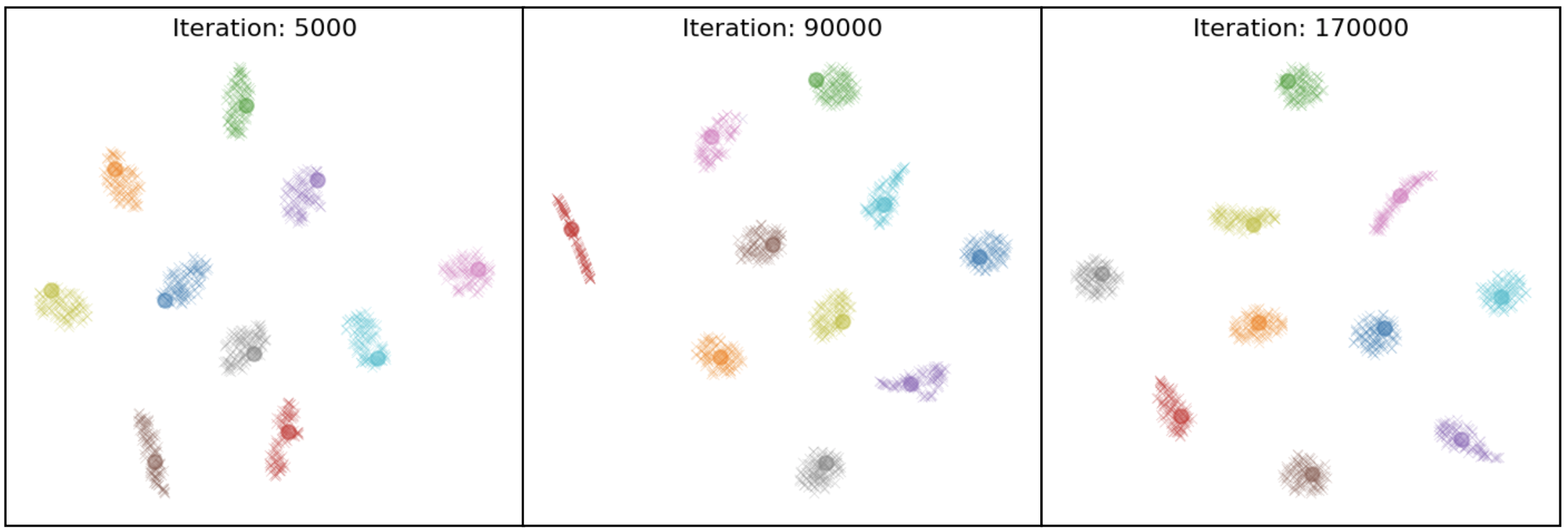}
    \caption{We randomly sample a data point from each category in the test set of CIFAR10. Then, 100 different adversarial samples corresponding to the natural images are generated. the latent representations of the natural and their adversarial counterparts are plotted using t-SNE. In the early stages of the robust training, the domain gap between the natural and adversarial samples is visible in the latent space. Later in the robust training, the adversarial samples start grouping around the natural sample such that it is harder for the generator to find adversarial directions.}
    \label{fig:tsne}
\end{figure*}

We also demonstrated the behavior of the latent space in Figure~\ref{fig:tsne}. In this figure, we randomly sample a data point from each category in the test set of CIFAR10. Similar to the above experiment, we generate 100 different adversarial samples corresponding to the natural image that is randomly chosen. Then, the latent representations of the natural and their adversarial counterparts are investigated using the t-SNE plot in Figure~\ref{fig:tsne}. In the early stages of the robust training, the domain gap between the natural and adversarial samples is visible in the latent space. As the model becomes more robust, it is harder for the generator to find adversarial directions. Thus, for the majority of the categories, the adversarial samples are grouped such that their center is the natural sample. 

\subsubsection{Ablation Study}
\label{subsect:ablation}
In this section, we investigate the necessity of the OT distance regularization and updating the weights of the generator with the OT distance rather than cross-entropy. In Table~\ref{tab:ablation}, \textbf{noReg + OT} denotes models trained without OT distance regularization, and \textbf{OT-Reg + Xent} represents models with the OT distance regularization in which the generator is updated with the cross-entropy loss. As can be observed in the table, OT distance regularization significantly improves the robustness of the model across all the datasets. In the presence of the regularizer, if the generator is updated with the cross-entropy loss, the adversarial accuracies of all three datasets decrease. This decrease is substantial for the SVHN dataset, which has completely different patterns than CIFAR10 datasets. 

\begin{table}[!t]
\centering
\caption{Ablation Results. \textbf{noReg + OT} denotes the model without OT distance regularization such that the generator is updated with OT distance between the natural and adversarial samples and the classifier is updated with only the cross-entropy loss. \textbf{OT-Reg + Xent} denotes the model with OT distance regularization where the generator is updated with the cross-entropy loss instead of the OT distance. Finally, \textbf{OT-Reg + OT} denotes the proposed model with OT distance regularization and the generator being updated with OT distance.}
\label{tab:ablation}
\centering
\begin{tabular}{|c|c|c|c|}
\multicolumn{4}{c}{\rule{0pt}{12pt}\textbf{CIFAR10}}                                                                                                                                                               \\ \hline 
\textbf{Defenses}                                                    & \textbf{Natural} &\textbf{PGD-20} &\textbf{CW-20} \\ \hline 
\textbf{noReg + OT} & 94.46 & 24.55 & 21.09\\ \hline
\textbf{OT-Reg + Xent} & 94.44 & 52.01 &36.70\\ \hline
\textbf{OT-Reg + OT (Proposed)} & 94.17 & 57.50 & 41.21\\ \hline

\multicolumn{4}{c}{\rule{0pt}{12pt}\textbf{CIFAR100}}                                                                                                                                                               \\ \hline 
\textbf{Defenses}                                                    & \textbf{Natural} &\textbf{PGD-20}&\textbf{CW-20}  \\ \hline 
\textbf{noReg + OT} & 77.40 & 26.33 & 8.98\\ \hline
\textbf{OT-Reg + Xent} & 76.95 & 42.28 & 30.32 \\ \hline 
\textbf{OT-Reg + OT (Proposed)} & 76.32 & 48.16 & 40.60\\ \hline

\multicolumn{4}{c}{\rule{0pt}{12pt}\textbf{SVHN}}                                                                                                                                                               \\ \hline 
\textbf{Defenses}                                                    & \textbf{Natural} & \textbf{PGD-20}&\textbf{CW-20}  \\ \hline 
\textbf{noReg + OT} & 96.06 & 6.66 & 7.06 \\ \hline
\textbf{OT-Reg + Xent} & 96.26 & 29.29 & 24.14 \\ \hline
\textbf{OT-Reg + OT (Proposed)} & 95.50 & 60.53 & 54.21\\ \hline
\end{tabular}
\end{table}

\section{Conclusion}
\label{sect:conc}
In this study, a robust training approach with generative adversarial perturbations is proposed. The attack generation does not require pixel to pixel translation or recurrent architectures. More importantly, the proposed generator does not take any gradient information as input. Thus, the computational complexity is reduced compared to the robust training models with iterative gradient computations. To encourage diverse adversarial perturbations during training, the attack generation network is updated by maximizing the optimal transport distance between the representations of the synthesized adversarial and natural samples. Furthermore, the perturbation is generated from a random vector. As a result, the dependency on a single sample and its label is reduced during the attack generation. The optimal transport distance between the adversarial and natural samples is also utilized to regularize the classifier such that the learned representations are encouraged to be robust against adversarial perturbations. Experiments on CIFAR10, CIFAR100, and SVHN datasets demonstrated that the proposed robust training approach can introduce adversarial robustness to the object recognition task without the degradation in the natural accuracy.

% or


%\appendices
%\section{}

\section*{Acknowledgment}

This work is supported by Bogazici University Research Fund under the Grant Number 17004.

\ifCLASSOPTIONcaptionsoff
  \newpage
\fi

\bibliographystyle{IEEEtran}
\bibliography{IEEEabrv,refs}


\begin{IEEEbiography}[{\includegraphics[width=1in,height=1.25in,clip,keepaspectratio]{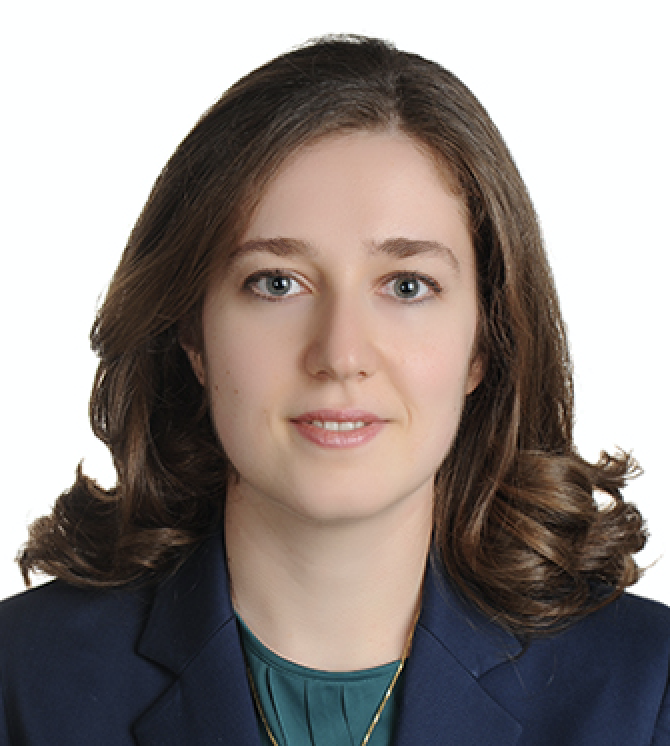}}]{\.{I}nci~M.~Bayta\c{s}}
is currently an Assistant Professor in the Department of Computer Engineering at Bo\u{g}azi\c{c}i University. She received her Ph.D. degree from the Department of Computer Science and Engineering at Michigan State University in 2019. Her research interests include machine learning, deep learning, adversarial robustness, temporal analysis, and biomedical informatics. She has served as program committee member for AAAI since 2019, and as reviewer for premier journals, such as IEEE Transactions on Knowledge and Data Engineering, IEEE Transactions on Pattern Analysis and Machine Intelligence, IEEE Journal of Biomedical and Health Informatics, and IEEE Transactions on Services Computing.
\end{IEEEbiography}

\begin{IEEEbiography}[{\includegraphics[width=1in,height=1.25in,clip,keepaspectratio]{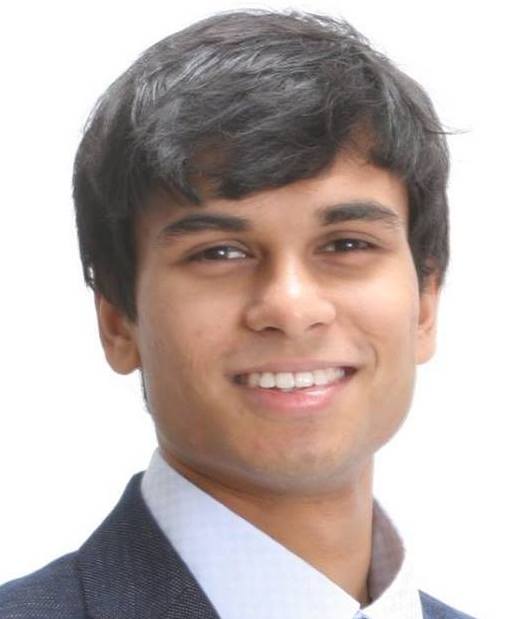}}]{Debayan Deb} He received his PhD degree in the Department of Computer Science and Engineering at Michigan State University in 2021. His research interests include pattern recognition, computer vision, and machine learning with applications in biometrics. He served as program committee member for CVPR and ICCV since 2020, as as reviewer for premier journals, including IEEE Transactions on Information Forensics \& Security and IEEE Transactions on Pattern Analysis and Machine Intelligence.
\end{IEEEbiography}


%\vfill


\end{document}